\documentclass[runningheads]{llncs}
\usepackage[T1]{fontenc}
\usepackage{booktabs}
\usepackage{amsmath}
\usepackage{epsfig}
\usepackage{url}
\usepackage{tabularx} % Load the tabularx package
\usepackage{lipsum} % Optional, for adding dummy text
\usepackage{caption}
\usepackage{longtable}
\usepackage{appendix}
\usepackage{multirow}
\usepackage{float}
\usepackage{mdframed}
\usepackage{graphicx}
\newcommand{\roberta}{RoBERTa}
\newcommand{\robertuito}{RoBERTuito}
\begin{document}
\title{Identification of emotions on Twitter during the 2022 electoral process in Colombia}
\author{Juan Jose Iguaran Fernandez\inst{1}\and
Juan Manuel Perez\inst{2}\and
Germán Rosati\inst{3}}
\authorrunning{Iguaran et al.}

\titlerunning{Identification of emotions on Twitter}
\institute{Maestría en Data Mining, Universidad de Buenos Aires (UBA), Argentina\\
\email{juanjose\_if3@hotmail.com}\\
\and
Instituto de Ciencias de la Computación, CONICET, Universidad de Buenos Aires,
Buenos Aires, Argentina\\
\email{jmperez@dc.uba.ar}\\
 \and
CONICET. Escuela IDAES, Universidad Nacional de San Martín, Argentina\\
\email{grosati@unsam.edu.ar}}
\maketitle              % typeset the header of the contribution
\begin{abstract}
    The study of Twitter as a means for analyzing social phenomena has gained interest in recent years due to the availability of large amounts of data in a relatively spontaneous environment. Within opinion-mining tasks, emotion detection is specially relevant, as it allows for the identification of people's subjective responses to different social events in a more granular way than traditional sentiment analysis based on polarity. In the particular case of political events, the analysis of emotions in social networks can provide valuable information on the perception of candidates, proposals, and other important aspects of the public debate. In spite of this importance, there are few studies on emotion detection in Spanish and, to the best of our knowledge, few resources are public for opinion mining in Colombian Spanish, highlighting the need for generating resources addressing the specific cultural characteristics of this variety.

    In this work, we present a small corpus of tweets in Spanish related to the 2022 Colombian presidential elections, manually labeled with emotions using a fine-grained taxonomy. We perform classification experiments using supervised state-of-the-art models (BERT models) and compare them with GPT-3.5 in few-shot learning settings. We make our dataset and code publicly available for research purposes.

\end{abstract}
\section{Introduction}

Twitter is a microblogging platform that can be examined as a forum for opinion analysis on social phenomena, especially in politics, despite its representativeness biases. Previous studies have explored how analyzing discussions on Twitter using NLP tools can reflect subjective perceptions of social phenomena \cite{o2010tweets}, \cite{tumasjan2010predicting}, \cite{mohammad2015sentiment}. However, there are few studies in Spanish for political events outside Iberian Spanish and, to the best of our knowledge, no resources are available for Latin American varieties of this language.

The evolution towards the use of neural networks, particularly the Transformer architecture \cite{vaswani2017attention}, has significantly improved the efficiency and accuracy of opinion mining tools by processing extensive texts and capturing the contextual complexity of language. This is particularly evident in BERT, a Transformer-based model pre-trained on large amounts of text, which is then fine-tuned for a specific task \cite{devlin2018bert}.

This study focuses on the analysis of emotions in tweets during the 2022 Colombian presidential elections using pre-trained language models. A total of 1,200 election-related tweets were collected, labeled, and used to fine-tune and evaluate pre-trained models. The dataset is offered as a resource for future research. Additionally, the same dataset was labeled using GPT-3.5, a large language model (LLM) created by OpenAI \footnote{\url{https://openai.com/}} from GPT-3 \cite{brown2020language}, with which one can interact with through text prompts. In this case, the task requested was the labeling of the tweets. The performance of these two labeling methods was compared.

This paper follows the following structure: Section \ref{sec:previous_works} provides an overview of the most relevant works related to this study. Section 3 details how the tweets were collected and the framework and results of the labeling process. Section 4 describes the classification experiments carried out. Section 5 shows the results that were obtained for both the fine-tuned model and GPT 3.5. Finally, section 6 provides some conclusions regarding this study.

\section{Previous works}
\label{sec:previous_works}

In this section we describe the most relevant studies to date related opinion mining and machine learning models used for this area. We also describe some of their applications in the context of social networks and social related phenomena.

\subsection{Supervised learning for emotions detection}

The automatic detection of emotional responses in text has long been a subject of interest. The significance of systems capable of identifying negative or positive opinions using online movie reviews has been highlighted in the past \cite{pang2002thumbs}. Additionally, the exploration of the semantic orientation of words to determine the overall polarity of texts has also been studied \cite{turney2002thumbs}.

To undertake the task of automatically detecting emotional states in text, it is usually necessary to have a labeled dataset from which automation can proceed. The creation of such resources has been undertaken in the past for individual words as well as more complex texts such as news articles or children's stories \cite{strapparava2004wordnet,wiebe2005annotating,alm2005emotions}. These resources later serve as training datasets for the task of supervised learning using traditional machine learning algorithms.

\subsection{Neural networks for text analysis}

Traditional techniques for analyzing emotions in text have focused on the relationship between terms and emotional states. Recently, the field has shifted to algorithms capable of capturing contextual relationships, such as neural networks \cite{acheampong2021transformer}.

Initially, the most popular architecture for NLP tasks was Recurrent Neural Networks (RNNs) \cite{cho2014learning}, due to their capacity to retain previous data outputs to predict new inputs. This architecture facilitated context analysis in text as it processes sequences of words. However, its sequential nature made it computationally expensive for handling large datasets. To address this issue, a new architecture called Transformers was introduced \cite{vaswani2017attention}. Transformers use a layer defined as self-attention for parallel processing, allowing for the consideration of each word's importance in context. This parallel processing capability enables the handling of large datasets, thus providing the model with sufficient training data to enhance language understanding.

Thanks to the parallelization and ability to retain distant word relationships in Transformers, highly predictive language models like BERT were developed \cite{devlin2018bert}. These models are trained on vast amounts of data, such as Wikipedia, to be pre-trained and then use contextual language representations for various NLP tasks by fine-tuning on specific tasks. For Spanish, a dedicated BERT model trained with diverse texts has been proposed \cite{canete2020spanish}, which outperforms multilingual BERT in Spanish evaluations.

\subsection{Sentiment and emotion on social networks}

The analysis of sentiment in online text, particularly on blogging platforms like Twitter, is crucial across various sectors including advertising, finance, and academia \cite{pang2008opinion}. Such platforms provide significant data for sentiment analysis \cite{aman2007identifying}, \cite{pak2010twitter}. In relation to social phenomena, Twitter sentiment has been found to correlate with opinion polls \cite{o2010tweets} and significant events \cite{bollen2011modeling}. In politics, Twitter sentiment often reflects public perception \cite{tumasjan2010predicting}.

S. M. Mohammad et al. extensively analyzed emotional content in tweets during the 2012 US presidential elections \cite{mohammad2015sentiment}. In their study, they developed a manual for labeling emotions which was provided to multiple labelers. These labelers were asked questions about the emotional content of a tweet and to select the emotion that best fit. Based on the resulting dataset, groups containing several emotions were created to provide a more robust dataset. From this dataset, a machine learning model was trained to identify emotions within tweets. This work has served as a primary influence on the present study.

Regarding Spanish resources for sentiment analysis and opinion mining in general, many resources have been created, particularly as part of workshops such as SEMEval and IberLEF. For instance, EmoEvent \cite{plaza2020emoevent} is a dataset for emotion detection in Spanish tweets related to several, distinct events. TASS \cite{garcia-vega2020overview} is a sentiment analysis dataset created in the shared task of the same name. To the best of our knowledge, there are no resources for emotion detection in Colombian Spanish tweets.

\section{Data}

This chapter discusses the data collection process carried out between the two electoral rounds, focusing on the utilization of political hashtags. It begins with an explanation of the filtering process aimed at retaining only relevant tweets, followed by an exploratory analysis of the data. Furthermore, it describes the labeling process, including the establishment of correlations between labels to form groups, and discusses the level of agreement among labelers.

\subsection{Data gathering}
The initial dataset comprises 585,001 tweets collected between May 22nd and June 22nd, 2022, during the presidential elections in Colombia. These tweets were extracted using 173 political trends, i.e., hashtags per day, obtained from websites that store historical trends \footnote{\url{https://getdaytrends.com/}} \footnote{\url{https://archive.twitter-trending.com/}} \footnote{\url{https://www.exportdata.io/trends/worldwide}}. After a filtering process, which removed tweets with fewer than 5 words, those with a proportion of mentions or hashtags exceeding 20% of the text, and those with links or from users with an atypical number of posts, the database was reduced to 193,348 tweets.

The hashtags were classified as Left, Right, or Neutral based on their content and perceived political trend. It is worth mentioning that even if a hashtag was associated with a particular political trend, it could still represent a different point of view. The distribution of hashtags was as follows: Neutral (40%), Left (28%), Right (28%) as shown in Figure \ref{figure:tweets_cantidad_hashtags}.

\begin{figure}[t]
    \centering
    \includegraphics[width=0.6\textwidth]{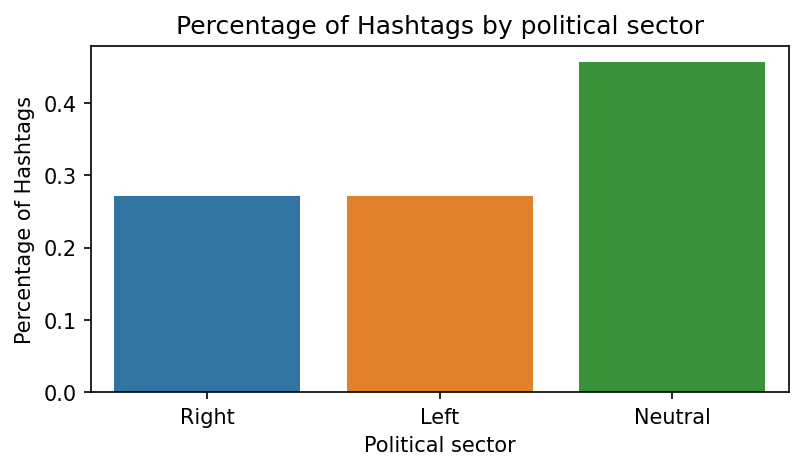}
    \caption{Percentage of Hashtags According to Assigned Political Orientation}
    \label{figure:tweets_cantidad_hashtags}
\end{figure}

\begin{figure}[!htbp]
    \centering
    \includegraphics[width=\textwidth]{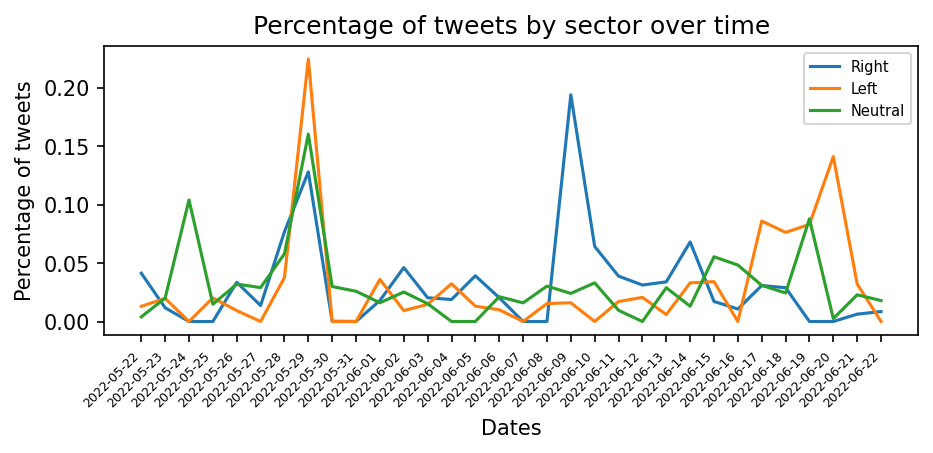}
    \caption{Percentage of tweets according to political orientation over time}
    \label{figure:tweets_porcentaje_tiempo}
\end{figure}

Temporal analysis reveals that certain dates, such as May 24th (the date of a debate), May 29th (the first round of elections), June 9th (an event known as the "Petro videos" where a leaked video regarding political strategy was made public), and dates around June 19th (the second round of elections), experienced peaks of activity. These peaks were observed across all three political sectors, as illustrated in Figure \ref{figure:tweets_porcentaje_tiempo}.

\subsection{Labeling}

A set of 1200 tweets was selected using stratified random sampling to preserve the original proportions of hashtags. This set was then subjected to manual classification by each of the authors, with labels corresponding to emotions identified in the tweets. Emotion, as defined by the APA \cite{vandenbos2007apa}, involves a complex reaction encompassing experiential, behavioral, and physiological elements. However, in this study, the focus was solely on the authors' expressive responses to labeling emotions, as subjective and physiological components were not accessible.

\begin{figure}[t]
    \centering
    \includegraphics[width=0.7\textwidth]{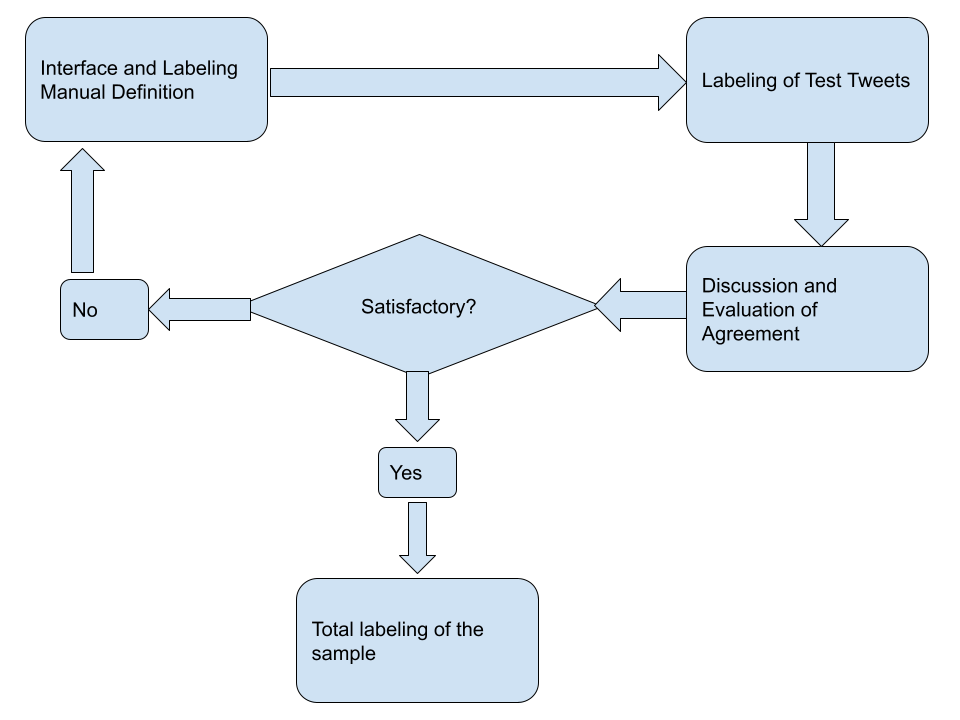}
    \caption{Annotation workflow}
    \label{figure:diagram}
\end{figure}

To establish a satisfactory labeling framework, an iterative process was developed following the methodology of Mohammad et al. \cite{mohammad2015sentiment}. A manual was created to outline the task and describe the possible labels. Subsequently, the authors labeled some tweets, measured the agreement of those labels, and engaged in discussions regarding the task's execution. This iterative process continued until the labeling and the output were deemed satisfactory. Details of this process are presented in Figure \ref{figure:diagram}.

\begin{figure}[t]
    \centering
    \includegraphics[width=\textwidth]{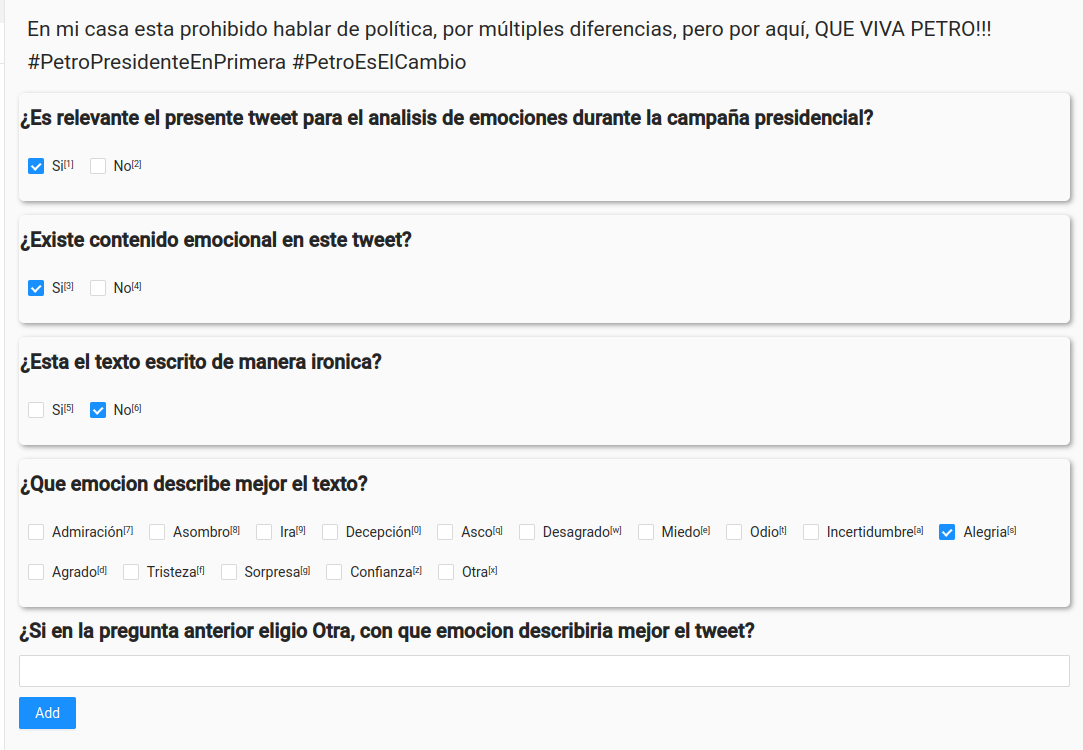}
    \caption{Labeling interface}
    \label{figure:interfaz}
\end{figure}

This process ultimately led to the development of a labeling interface using the web platform Label Studio \footnote{\url{https://labelstud.io/}}. The interface enabled the assignment of one or multiple emotions to each tweet via a multiple-choice scheme featuring 14 emotions and an "Other" category. Prior to labeling, participants were prompted to determine if the tweet contained emotional content. The resulting interface is illustrated in Figure \ref{figure:interfaz}.

\begin{figure}[t]
    \centering
    \includegraphics[width=\textwidth]{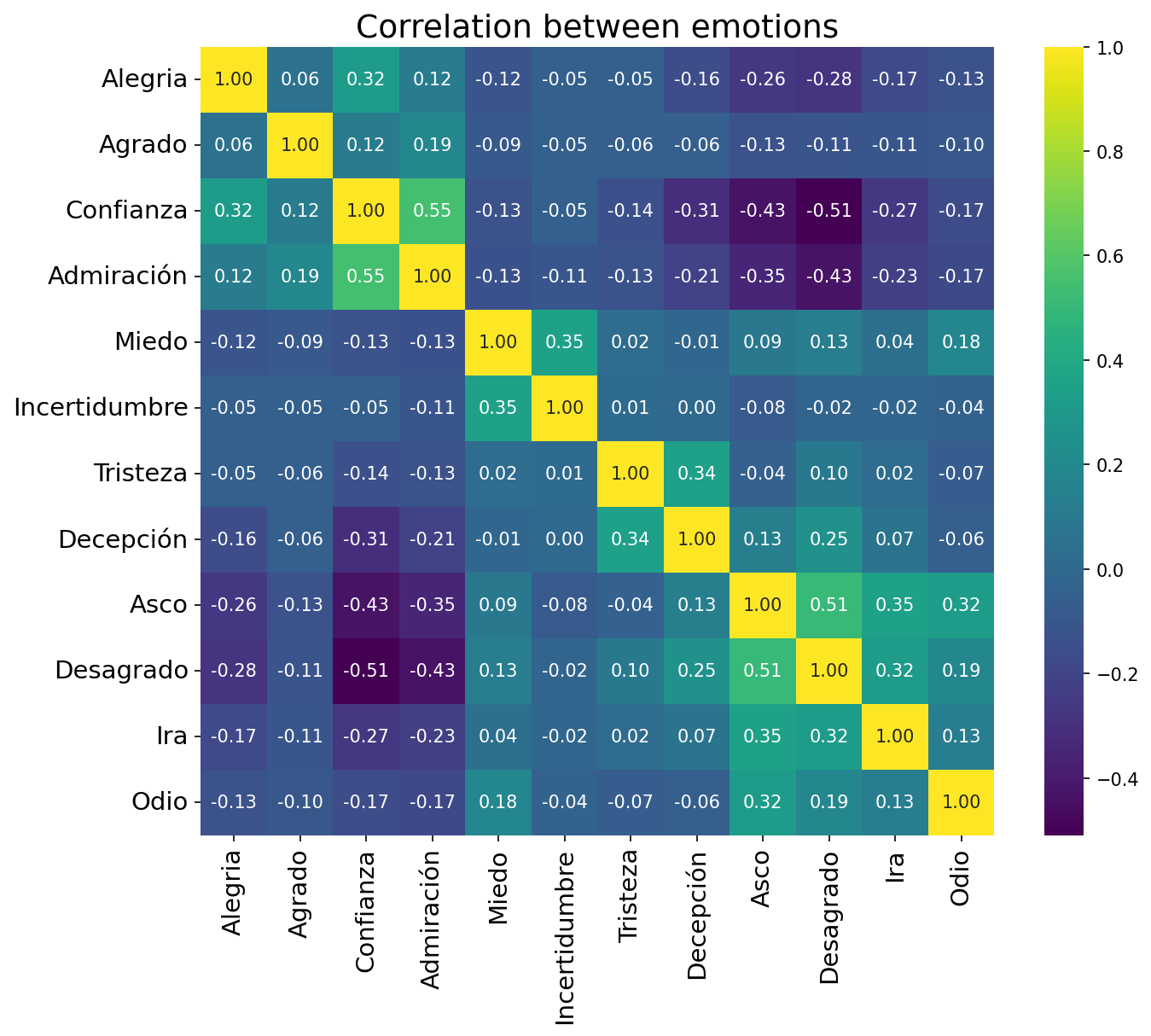}
    \caption{Correlation index between emotion labels assigned to the tweets}
    \label{figure:correlacion_emociones}
\end{figure}

After labeling, a database containing the assigned labels for each tweet by each author was generated. This database was then utilized to calculate the correlation between labels assigned by the authors. It was observed that labelers assigned semantically similar labels to certain tweets, as depicted in Figure \ref{figure:correlacion_emociones}.

Based on these results, labels were grouped into four categories based on their correlation: Joy, Fear, Sadness, and Disgust. Final labels were assigned if at least two labelers assigned some of the granular emotions that make up a specific group. The Fleiss Kappa index was then used to measure agreement among labelers, with results shown in Table \ref{table:agreements}.

\begin{table}[b]
    \centering
    \caption{Fleiss Kappa index for each emotion}
    \label{table:agreements}
    \begin{tabular}{lllll}
        \toprule
                         & alegria & miedo & tristeza & asco \\
        \midrule
        Number of Tweets & 464     & 98    & 103      & 580  \\
        Fleiss k index   & 0.69    & 0.47  & 0.4      & 0.62 \\
        \bottomrule
    \end{tabular}
\end{table}

\begin{figure}[t]
    \centering 	\includegraphics[width=0.5\textwidth]{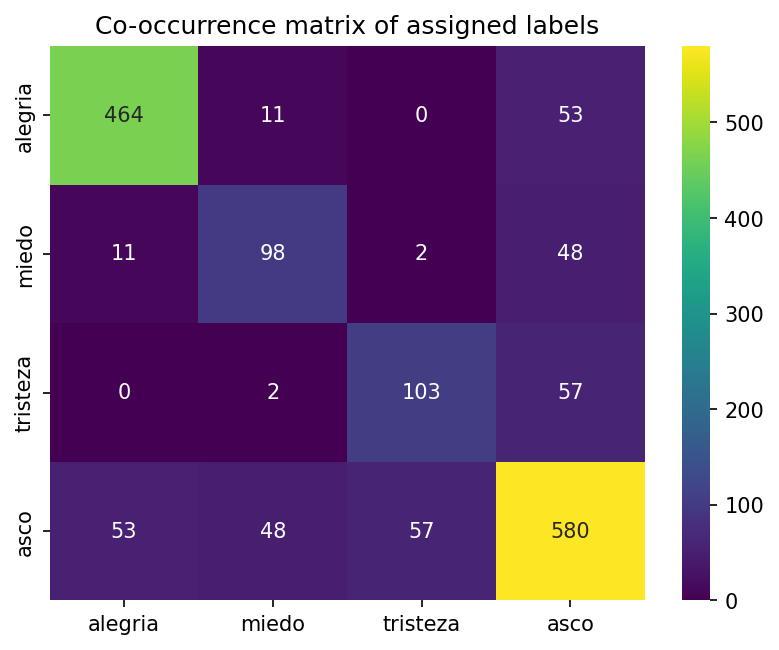}
    \caption{Co-occurrence matrix between assigned labels}
    \label{figure:Matriz_etiquetas}
\end{figure}

Joy and Disgust exhibited higher scores compared to Sadness and Fear. It is noteworthy that the emotions with better performance also had a higher number of tweets. Additionally, during the labeling process, overlap was observed between fear and disgust, as well as between sadness and disgust, leading to instances where only one labeler identified one of these emotions, resulting in disagreement. The final number of tweets labeled for each emotion, as well as the overlap between them, is illustrated in Figure \ref{figure:Matriz_etiquetas}. The final dataset containing the labels assigned for each tweet is available for the community \footnote{\url{https://huggingface.co/datasets/jjiguaran/tweets_emotions_elections_colombia}}.

\section{Classification experiments}

In this section, two approaches for classifying the dataset are explained: pre-trained models and the Large Language Model (LLM), detailing their training process and evaluation metrics.

\subsection{Pre-trained models fine-tuning}

Pretrained language models are accessible on the Hugging Face platform \footnote{\url{https://huggingface.co/}}, utilizing the Transformers library \footnote{\url{https://huggingface.co/docs/transformers/index}}. For fine-tuning, three pre-trained language models were chosen: \robertuito{} \cite{perez-etal-2022-robertuito}, BETO \cite{CaneteCFP2020}, and \roberta{} \cite{ROBERTA}. These models were selected because they were specifically trained for Spanish, and in the case of \robertuito{}, for the Twitter context.

These models underwent training and evaluation using K-fold cross-validation, where the dataset was partitioned into train and test sets k times. The model was trained on the train partition and evaluated on the test set. In this case, k was set to 5.

The performance of the model was assessed using the F1 score metric, defined as follows:

\begin{equation}
    {\text{F1}}=\frac{\text{TP} }{\text{TP} + \frac{1}{2}\text{(FP +FN)}}
\end{equation}

Where TP represents the number of true positives, FP stands for the number of false positives, and FN indicates the number of false negatives across all classes. Similarly, the Micro F1 score is calculated, but it considers all present classes. This metric was chosen because the model allows for multiple classifications, enabling the simultaneous evaluation of the performance of different classes.

The main hyperparameters employed include the implementation of AdamW, an optimization algorithm proposed by \cite{loshchilov2017decoupled}. A learning rate of 5e-05 was utilized. Additionally, 3 training epochs were conducted on the training set, with a batch size of 8.

\subsection{Few-shot learning}

The few-shot approach was conducted using OpenAI's API \footnote{\url{https://openai.com/blog/openai-api}} to establish a connection to their GPT-3.5 model. Once connected, a prompt is submitted to the model, defining a request. In this case, the prompt requested the model to label the provided tweet with one of the four emotions and to label it as "other" if it did not match any of them. A brief description of the emotion was included within the prompt, along with a few examples of tweets already classified with their respective labels. The model was also asked to provide an explanation of how it arrived at the given label.

The prompt we used was as follows:

\begin{quote}
    Asigna una o varias de siguientes las emociones segun correspondan
    al tweet dentro de la siguiente lista:

    - Alegria: Positividad, entusiasmo, apoyo, confianza, celebracion o
    gratitud.

    - Tristeza: El tweet expresa emociones de dolor emocional, desanimo,
    decepcion o pesar

    - Asco: Expresiones intensas de aversion, desprecio o ataques nega-
    tivos.
    - Miedo: Sensacion de amenaza, inquietud, incertidumbre o ansiedad

    Los tweets a analizar estan en el marco de las elecciones colombianas
    del año 2022. Si un tweet no encaja con ninguna emocion descrita o el
    contenido emocional no es suficientemente evidente, etiquetalo como
    "Otra".

    Describe paso a paso el razonamiento que tuviste para llegar a esas emociones.
\end{quote}

\section{Results}

Table \ref{table:models_performance} displays the results obtained by each model evaluated in the dataset. It shows that, among the supervised algorithms, the best-performing models are \roberta{} and \robertuito{}, both of which exhibit higher performance for joy and disgust compared to fear and sadness. This aligns with the agreement scores observed during the labeling process, indicating that the overlap between disgust and fear/sadness may pose a challenge for the model.

\begin{table}[t]
    \centering
    \caption{Performance metrics of models}
    \label{table:models_performance}
    \begin{tabular}{p{2cm} c c c c c}
        \hline
        Model         & Joy             & Disgust         & Fear            & Sadness         & Micro F1 \\
        \hline
        \roberta{}    & $81.2$          & $\mathbf{82.5}$ & $28.1$          & $35.6$          & $73.4$   \\
        BETO          & $78.1$          & $77.8$          & $31.3$          & $32.4$          & $68.3$   \\
        \robertuito{} & $82.0$          & $80.5$          & $35.8$          & $40.6$          & $72.9$   \\
        GPT-3.5       & $\mathbf{84.1}$ & $78.8$          & $\mathbf{52.0}$ & $\mathbf{45.5}$ & $75.2$   \\
        \hline
    \end{tabular}
\end{table}

We can observe that in the case of GPT-3.5 the best-performing labels were joy and disgust, while fear and sadness performed the worst. However, it is noticeable that GPT's performance is considerably better than that of the trained models for these emotions. The performance for joy is slightly better and for disgust slightly worse compared to \roberta{} and \robertuito{}. This demonstrates that the Large Language Model (LLM) seems to perform better when distinguishing between negative emotions, and is quite better for underrepresented emotions such as fear and sadness.

To have a better understanding of the weak points of GPT-3.5, we conducted an error analysis. Table \ref{tab:error_analysis} shows some examples of the errors made by the model. While this task is highly subjective and ---in some cases--- disagreement might not necessarily mean an error, in some cases the predictions show a clear misunderstanding of the tweet. For this specific task, we can observe that, in some cases, the model is struggling with some aspects of ironic/sarcastic tweets or not identifying the use of some slang words or expressions. In other selected examples, GPT-3.5 fails to detect mixed emotions in tweets (e.g. both disgust and joy).

\setlength{\tabcolsep}{3pt} % Adjust the 10pt value as needed

\begin{table}[htbp!]
    %\captionsetup{font=scriptsize} % Set caption font size
    \footnotesize
    \centering

    \begin{tabular}{p{5.25cm} p{2cm} p{4.5cm}}
        \hline
        \textbf{Text}                                                                                                                                                                                                                             & \textbf{Labels}  & \textbf{CoT Prediction}                                                                                                                                                                                        \\
        \hline
        \multicolumn{3}{c}{Irony not detected}                                                                                                                                                                                                                                                                                                                                                                                                                                        \\
        \hline
        %When Petro is interviewed, you learn about History, Macroeconomics, Green Policies and when Fico is interviewed, I'm checking if they stole my cell phone, my wallet, this guy is scary guy....                                             & Joy, disgust    & The author expresses his support and admiration for Petro, highlighting his qualities and knowledge. On the other hand, he expresses contempt and fear towards Fico, considering him a threat. The final response is: Joy, Disgust, Fear \\

        \\
        @lcvelez @lafm an alternative to exercise the right to be informed as voters, don't hide the old man, don't feel sorry for them. @petrogustavo is ready.                                                                                  & Disgust          & The author expresses his support towards the idea of a mandatory debate to exercise the right to be informed and criticizes the attitude of hiding the older candidate. The final response is: joy.            \\

        From very early in the morning supporting Marbelle in her dream of leaving the country.                                                                                                                                                   & Disgust          & The author expresses his support and enthusiasm for Marbelle in her decision to leave the country, which reflects positivity and celebration. The final response is: joy.                                      \\
        As it is already known that the Retrograde Rodolfo Hernandez is going to win by order of the great mafia elites of Colombia, so let the great mafia elites of Colombia tremble, because he, Rodolfo, is going to put an end to CORRUPTION & Disgust          & The author expresses his confidence in the victory of Rodolfo Hernandez and his contempt for the mafia elites, and also shows support for the fight against corruption. The final answer is: joy, disgust.     \\
        \hline
        \multicolumn{3}{c}{Missing mixed emotion}                                                                                                                                                                                                                                                                                                                                                                                                                                     \\
        \hline
        It is not objective. It seems to me that the moderators are biased. I expected a little more from @NoticiasCaracol but well...What else is expected from a private Colombian media.                                                       & Sadness, Disgust & The author expresses his dissatisfaction with what he considers a biased debate, showing his disappointment and distrust towards the moderators and the Colombian private media. The final answer is: sadness. \\
        ElQueLosDELATA Tomorrow we will go to the polls to defeat this mafia!                                                                                                                                                                     & Joy, disgust     & The author expresses enthusiasm and support towards the next day's election day, with the goal of defeating those he considers a gang of mafiosi. The final response is: joy.                                  \\
        \hline
    \end{tabular}
    \caption{Error analysis for GPT-3.5 Chain-of-Thought (CoT) predictions. The table shows some examples where the model fails to detect irony or mixed emotions.}
    \label{tab:error_analysis}
\end{table}
\setlength{\tabcolsep}{6pt} % Adjust the 10pt value as needed

\section{Conclusions}

This study aimed to compare the performance of fine-tuned pre-trained language models, namely \robertuito{}, RoBERTa, and BETO, against the Large Language Model (LLM) GPT-3.5 through few-shot learning in identifying emotions present in tweets related to the 2022 presidential elections in Colombia.

This comparison was facilitated by the creation of a dataset of 1200 tweets manually labeled by the authors. The labeling task utilized a web interface and followed an internal manual, allowing the assignment of one or several of the 14 available emotions to each tweet. To assign an emotion to a tweet, at least two annotators had to agree on it. Finally, each tweet was classified with one of the four labels resulting from grouping the original labels based on their correlation. It is important to note the inherent difficulty of the labeling task, as it aimed to achieve an objective classification of a subjective activity, necessitating an iterative process to develop a manual and a labeling interface that approached this purpose. Additionally, although all annotators are native Spanish speakers, only the author is originally from Colombia, which led to certain language usages or specific contextual situations being clearer to him than to the other annotators.

The results of the models revealed a much greater presence of joy and disgust than fear and sadness, consistent with observations during the labeling process, where these two emotions were less frequent and often accompanied by the emotion of disgust. These factors also explain the relatively lower predictive capacity for less prevalent emotions compared to joy and disgust.

It is noteworthy that GPT-3.5 significantly outperformed the fine-tuned models in the emotions of fear and sadness. This demonstrates the LLM's capability to perform as well as fine-tuned models for the most prevalent emotions, as well as its ability to excel in distinguishing negative emotions. Also, GPT-3.5 seems to be struggling with some aspects of irony and sarcastic tweets.

\section{Limitations}

One of the main limitations of this study is the small size of the dataset and the short period of time in which the tweets were collected. This has led to a lack of diversity in the data, which is somehow reflected in the distribution of perceived emotions. Additionally, the labeling process was performed by native Spanish speakers, but only one of them was Colombian, which could have led to some misunderstandings in the interpretation of the tweets.

Finally, the comparison between the fine-tuned models and GPT-3.5 was not exhaustive, as the latter was only used in few-shot learning without trying several prompts. This might have led to suboptimal performance for the LLMs.

\end{document}